\documentclass{article}

\usepackage{emnlp2017}
\emnlpfinalcopy
\usepackage{times}
\usepackage{latexsym}

\usepackage{graphicx}
\usepackage{subfigure}

\usepackage{natbib}

\usepackage{tikz}
\usepackage{pgfplots}
\usetikzlibrary{arrows.meta}

\usepackage{algorithm}
\usepackage{algorithmic}

\usepackage{amsmath}
\usepackage{bm}
\usepackage{stmaryrd}  

\DeclareMathOperator*{\argmax}{arg\,max}

\usepackage{hyperref}

\title{Multiscale sequence modeling with a learned dictionary}

\author{Bart van Merri\"enboer \\ MILA \And
    Amartya Sanyal \\ IIT Kanpur \And
    Hugo Larochelle \\ Google Brain \And
    Yoshua Bengio \\ MILA, CIFAR Fellow}

\begin{document}

\maketitle

\begin{abstract}
We propose a generalization of neural network sequence models. Instead of predicting one symbol at a time, our multi-scale model makes predictions over multiple, potentially overlapping multi-symbol tokens. A variation of the byte-pair encoding (BPE) compression algorithm is used to learn the dictionary of tokens that the model is trained with. When applied to language modelling, our model has the flexibility of character-level models while maintaining many of the performance benefits of word-level models. Our experiments show that this model performs better than a regular LSTM on language modeling tasks, especially for smaller models.
\end{abstract}

\section{Introduction}

\begin{figure*}[h]
  \centering
\begin{tikzpicture}[-latex,node distance=2cm]
\tikzstyle{state} = [circle,draw,minimum width=1cm]
\node (a)[state] {$\varepsilon$};
\node (b)[state,right of=a] {H};
\node (c)[state,right of=b] {He};
\node (d)[state,right of=c] {Hel};
\node (e)[state,right of=d] {Hell};
\node (f)[state,right of=e] {Hello};
\node (g)[state,right of=f] {Hello$\cdot$};
\tikzstyle{nonstate} = [circle,draw,dotted,minimum width=1cm]
\node (b1)[nonstate,above of=b] {G};
\node (c2)[nonstate,below of=c] {Hf};
\node (d1)[nonstate,above of=d] {Hek};
\node (d2)[nonstate,below of=d] {Hem};
\node (e1)[nonstate,above of=e] {Helk};
\node (f2)[nonstate,below of=f] {Hellp};
\draw (a) -- node[above] {H} (b);
\draw (b) -- node[above] {e} (c);
\draw (c) -- node[above] {l} (d);
\draw (d) -- node[above] {l} (e);
\draw (e) -- node[above] {o} (f);
\draw (f) -- node[above] {$\cdot$} (g);
\tikzstyle{trans1} = [dotted,bend left,below right]
\tikzstyle{trans2} = [dotted,bend right,above right]
\draw (a) edge [trans1] node[trans1] {G} (b1);
\draw (b) edge [trans2] node[trans2] {f} (c2);
\draw (c) edge [trans1] node[trans1] {k} (d1);
\draw (c) edge [trans2] node[trans2] {m} (d2);
\draw (d) edge [trans1] node[trans1] {k} (e1);
\draw (e) edge [trans2] node[trans2] {p} (f2);
\end{tikzpicture}
  \caption{\label{fig:charmc} Diagrammatic representation of a character-level language model as a Markov chain. The probability of the string ``Hello'' is the probability of reaching the absorbing state ``Hello$\cdot$'' starting from the empty string ($\varepsilon$), where $\cdot$ is a special end-of-string (EOS) token. Each state transition is analogous to concatenating a token from the dictionary $\Sigma$ to the state.}
\end{figure*}
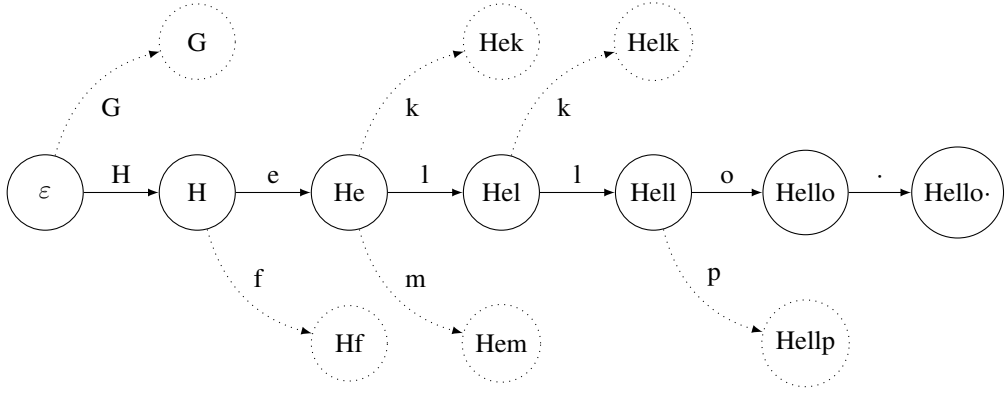

Sequence modeling is the task of learning a probability distribution over a set of finite sequences. We consider sequences of elements drawn from a finite set of symbols, $s_t \in \Sigma$. In the context of language modeling this is the probability mass function of a set of strings given some alphabet of characters.
\begin{equation}
p\left(s_1\dots s_n\right),\quad s_t \in \Sigma
\end{equation}
Most approaches in language modeling follow~\cite{Shannon-1948}
and model sequences as acyclic Markov chains, exploiting the fact that natural languages have strong temporal dependencies (see figure \ref{fig:charmc}). 
\begin{equation}
\label{eq:prod}
p\left(s_{1}\dots s_{n}\right)\approx \prod_{t=1}^{n}p\left(s_{t}\mid s_{1}\dots s_{t-1}\right),\quad s_{t}\in \Sigma
\end{equation}

Recurrent neural networks (RNNs) can be used to efficiently model these Markov chains~\cite{Mikolov-thesis-2012,mikolov2010recurrent}. The hidden state of the network can encode the subsequence that is conditioned on ($s_1\dots s_{t-1}$) using constant memory. Let $\bm{x}_t$ be an embedding of symbol $s_t$, then an RNN model is of the form
\begin{align}
\bm{h}_t &= f(\bm{x}_t,\bm{h}_{t-1}) \\
\bm{y}_t &= g(\bm{h}_t)
\end{align}

Typically the function $f$ is a long-short term memory (LSTM) unit or gated recurrent unit (GRU), and $g$ is a linear transformation followed by a softmax activation.


\subsection{Tokenization}

\begin{figure*}[h]
  \centering
\begin{tikzpicture}[-latex,node distance=2cm]
\tikzstyle{state} = [circle,draw,minimum width=1cm]
\node (a)[state] {$\varepsilon$};
\node (b)[right of=a] {};
\node (c)[right of=b] {};
\node (d)[state,right of=c] {Hello};
\node (e)[right of=d] {};
\node (f)[right of=e] {};
\node (g)[state,right of=f] {Hello$\cdot$};
\tikzstyle{nonstate} = [circle,draw,dotted,minimum width=1cm,align=center]
\node (d1)[nonstate,above of=d] {Hallo};
\node (d2)[nonstate,below of=d] {Hi};
\node (g1)[nonstate,above of=g] {Hello\\world};
\node (g2)[nonstate,below of=g] {Hello\\there};
\draw (a) -- node[above] {Hello} (d);
\draw (d) -- node[above] {$\cdot$} (g);
\tikzstyle{trans1} = [dotted,bend left,below right]
\tikzstyle{trans2} = [dotted,bend right,above right]
\draw (a) edge [trans1] node[trans1] {Hallo} (d1);
\draw (a) edge [trans2] node[trans2] {Hi} (d2);
\draw (d) edge [trans1] node[trans1] {world} (g1);
\draw (d) edge [trans2] node[trans2] {there} (g2);
\end{tikzpicture}
  \caption{\label{fig:wordmc} A word-level language model, requiring fewer transitions in order to reach the state ``Hello$\cdot$''. The state space is significantly reduced, which means that many strings cannot be modeled.}
\end{figure*}
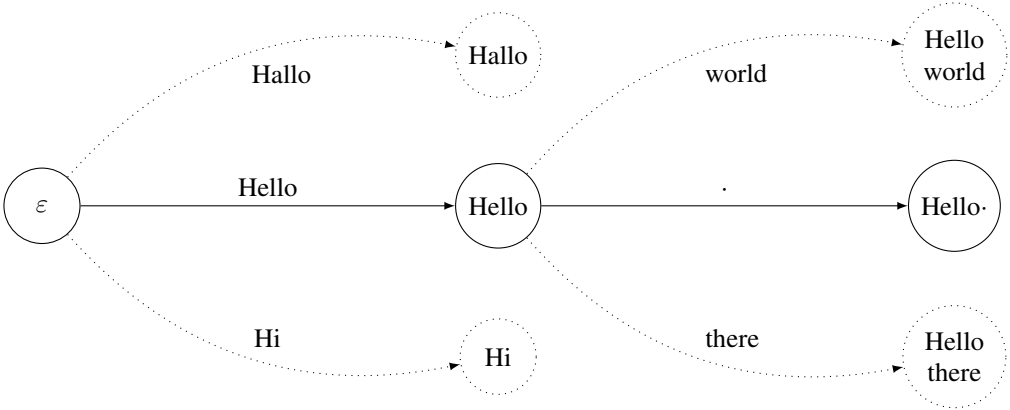

Natural language is naturally represented as a sequence of characters~\cite{Sutskever-et-al-ICML2011,Mikolov-thesis-2012,Graves-arxiv2013}. However, in practice text is usually `tokenized' and modeled as a sequence of words instead of characters (see figure~\ref{fig:wordmc}). Word-level models display superior performance to character-level models, which we argue can be explained by several factors.

\subsubsection{Training difficulties}

Tokenization reduces the length of the sequences to model. Learning long-term dependencies with RNNs can be difficult~\cite{Pascanu-et-al-ICML2013,Bengio-et-al-TNN1994,Hochreiter91}, and in natural language dependencies such as agreement in number or case can span tens or even hundreds of characters.

Furthermore, the softmax generally used in neural networks can never assign a probability of exactly 1 to the correct token. The product of many probabilities less than $1$ in equation~\ref{eq:prod} causes the probability of a sequence to decay quickly as it grows longer. To counteract this behaviour a network will quickly learn to fully saturate the softmax. However, this slows down learning~\cite{LeCun+98backprop}.

\subsubsection{Compositionality}
\label{sec:compositionality}

In the context of natural language it can be argued that character-level and word-level language modeling are fundamentally different.

Word-level models rely on the principle of compositionality~\cite{sep-compositionality} and can learn to represent the meaning of a sentence by combining the semantic representations of its constituent words~\cite{mitchell2008vector}. This mapping is arguably `smooth': If words have similar semantics, the sentences they form are likely to have a similar meaning and representation as well.

A character-level model performs a second, different task: Mapping a sequence of characters to the semantic representation of a morpheme. The principle of compositionality does not apply in this case. It is a lookup operation which is entirely non-linear: `the', `then' and `they' are entirely unrelated. It is possible that character-level RNN models perform worse than word-level models because RNNs are ill-suited to perform this lookup operation. It is feasible that other application domains have a similar hierarchical structure in their sequential data e.g. sequence motifs in genetics.

\section{Multi-scale sequence modeling}

\label{sec:multi}

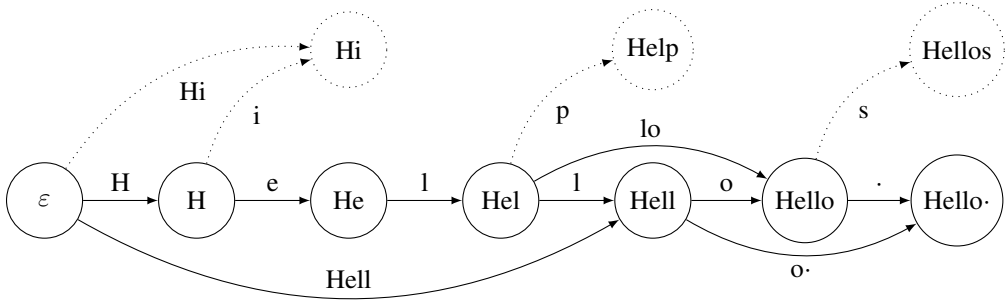
\begin{figure*}[h]
  \centering
  \begin{tikzpicture}[-latex,node distance=2cm]
\tikzstyle{state} = [circle,draw,minimum width=1cm]
\node (a)[state] {$\varepsilon$};
\node (b)[state,right of=a] {H};
\node (c)[state,right of=b] {He};
\node (d)[state,right of=c] {Hel};
\node (e)[state,right of=d] {Hell};
\node (f)[state,right of=e] {Hello};
\node (g)[state,right of=f] {Hello$\cdot$};
\tikzstyle{nonstate} = [circle,draw,dotted,minimum width=1cm]
\node (c1)[nonstate,above of=c] {Hi};
\node (e1)[nonstate,above of=e] {Help};
\node (g1)[nonstate,above of=g] {Hellos};
\draw (a) -- node[above] {H} (b);
\draw (b) -- node[above] {e} (c);
\draw (c) -- node[above] {l} (d);
\draw (d) -- node[above] {l} (e);
\draw (a) edge[bend right] node[above] {Hell} (e);
\draw (e) -- node[above] {o} (f);
\draw (e) edge[bend right] node[below] {o$\cdot$} (g);
\draw (d) edge[bend left] node[above] {lo} (f);
\draw (f) -- node[above] {$\cdot$} (g);
\tikzstyle{trans1} = [dotted,bend left,below right]
\tikzstyle{trans2} = [dotted,bend right,above right]
\draw (a) edge [trans1] node[trans1] {Hi} (c1);
\draw (b) edge [trans1] node[trans1] {i} (c1);
\draw (d) edge [trans1] node[trans1] {p} (e1);
\draw (f) edge [trans1] node[trans1] {s} (g1);
\end{tikzpicture}
  \caption{\label{fig:multimc}The multi-scale model allows multiple outgoing transitions, maintaining the flexibility of a character-level model while incorporating many of the benefits of word-level models. Any path through the Markov chain from $\varepsilon$ to Hello$\cdot$ is a segmentation of the string Hello using the tokens in the dictionary. The probability of the state Hello$\cdot$ is the sum of the likelihood of each segmentation. When modeled using an RNN, each state corresponds to a hidden state $\bm{h}_t$, and each arrow corresponds to the application of the transition function $f$ which takes inputs $\bm{h}_t$ and token embedding $\bm{x}_i$.}
\end{figure*}

In typical sequence models, we model the likelihood of the next symbol individually. A single symbol (e.g. a character) is selected from a dictionary of mutually exclusive options. In this paper we propose a more general setting in which at each step we make predictions over multi-symbol tokens, potentially multiple of which are correct if they share a prefix (see figure~\ref{fig:multimc}).

Formally, given a set of symbols $\Sigma$, consider a dictionary of multi-symbol tokens $T$, where $\Sigma \subset T$. (This condition guarantees that the space of sequences we can model is the same as for typical symbol-level models.) Let $|t_i|$ denote the number of symbols in token $t_i$. The Markov chain (see figure~\ref{fig:multimc}) for a sequence $s_1\dots s_n$, $s_t \in \Sigma$, can be modeled using an RNN as follows:
\begin{align}
\bm{h}_t &= \frac{1}{|T_t|}\sum_{T_t} f(\bm{x}_i,\bm{h}_{t-|t_i|}), \label{eq:average} \\
&\quad T_t = \{t_i : t_i \in T, t_i = s_{t-|t_i|+1}\dots s_t\} \\
\bm{y}_t &= g(\bm{h}_t)
\end{align}
where $\bm{x}_i$ is an embedding of token $t_i$. Note that a typical RNN model (e.g. a character-level language model) is a special case of this model where $T = \Sigma$.

The likelihood of this model is tractable and can be easily calculated using dynamic programming. We can optimize this likelihood directly using gradient descent. This is similar to the forward-backward algorithm used in hidden Markov models and connectionist temporal classification (CTC)~\cite{Graves-et-al-2006}.

\begin{align}
p\left(s_{1}\dots s_{t}\right) = \sum_{T_t} p\left(t_i|s_{1}\dots s_{t-|t_i|}\right) p\left(s_{1}\dots s_{t-|t_i|}\right)
\end{align}

This approach can be used in general for the modeling of Markov chains without cycles in the case of a finite set of transitions (even if the state space is infinite). The recurrent neural network predicts the transition probabilities over this finite set of transitions for each state using a representation of the state and a learned representation of each transition. We believe this is a novel approach to modelling acyclical Markov chains using RNNs.

In this work we consider a multiscale generalization of LSTM networks. For transition functions, $f$, with multiple operations we can perform the averaging at any point. We choose to average the cell states and output gates. Note that performing the averaging earlier on reduces the amount of computation.

\[
\begin{pmatrix}
\bm{f_i} \\
\bm{i_i} \\
\bm{o_i} \\
\bm{g_i} \end{pmatrix} = \bm{W}_h \bm{h}_{t - |t_i|} + \bm{W}_x \bm{x}_i + \bm{b}
\]
\[
\bm{c}_t = \frac{1}{N} \sum_{i = 1}^{N} \left(\sigma(\bm{f}_i) \odot \bm{c}_{t - |t_i|} + \sigma(\bm{i}_i) \odot \tanh(\bm{g}_i) \right)
\]
\[
\bm{h}_t = \sigma(\frac{1}{N}\sum_{i=1}^{N}\bm{o}_i) \odot \tanh(\bm{c}_t)
\]

\subsection{Model characteristics}

\label{sec:characteristics}

The computational complexity of a regular RNN model grows as a function of the sequence length, $O(T)$. The multiscale model's complexity instead grows as a function of the number of arcs. The number of arcs in a sequence is theoretically bounded by $\frac{T(T + 1)}{2}$, but in practice it grows sublinearly with the size of the dictionary. For example, for a dictionary with 16384 tokens we find an average of 2.7 arcs per time step for the text8 dataset.

It should be noted that the computation of arcs can be entirely parallelized, so on a parallel computer (e.g. a GPU) the span (depth) of the computation is equivalent to that of a normal RNN, $O(T)$.

During training time the memory usage of an RNN model grows as $O(T)$ because of the need to store the hidden states for the backward propagation. The multiscale model's memory usage grows the same and does not depend on the number of arcs, since the averages (see formula \ref{eq:average}) can be calculated by accumulating values in-place. The need to keep token embeddings in memory means that the memory usage grows as $O(T + D)$ where $D$ is the dictionary size.

In conclusion, the multiscale model is both computationally and memory efficient. On a parallel architecture it has a the same computational complexity as a regular RNN and only requires a small amount of extra memory in order to store the token embeddings.

\subsection{Dictionary learning}

The formulation of our multi-scale model requires the construction of a dictionary of multi-symbol tokens. Heuristically speaking, we would simplify our modeling problem if we construct a dictionary which allows each sequence to be segmented into a short sequence of tokens, minimizing the shortest path length through the graph (see figure~\ref{fig:multimc}).

In natural language processing, word-level models usually construct a dictionary by splitting strings on whitespace and punctuation. The dictionary then consists of the $N$ most frequent tokens, with the rest of the words replaced with a special out-of-vocabulary (OOV) token. Note that many other application domains (e.g. modeling DNA sequences) don't have any straightforward heuristics to tokenize the data.

Even in language modeling this type of tokenization is problematic for a variety of reasons. The number of words in natural language is effectively infinite for synthetic languages, which means there will always be OOV tokens. Furthermore, it is arguably arbitrary from a linguistic perspective. Whereas English is a rather isolating language, with $\sim$1.67 morphemes per word~\cite{greenberg1960quantitative} on average, synthetic languages such as Turkish or Eskimo have $\sim$2.33 and $\sim$3.70 morphemes per word respectively. For example, the Dutch word \emph{meervoudigepersoonlijkheidsstoornis} (multiple personality disorder) contains 10 morphemes. For these types of languages, we might want to consider a tokenization that contains subword units. On the other hand, for highly isolating languages we might want to model several words as a single token e.g. \emph{chúng tôi}, Vietnamese for `we'.

\subsection{Dictionary coders}



Instead of arbitrarily splitting on whitespace, a more principled approach is to to `learn' the tokens to be modeled. Here we propose an approach which is grounded in text compression and inspired by the byte-pair encoding (BPE) algorithm. BPE has been used in the domain of neural machine translation to learn subword units, reducing the number of OOV tokens~\cite{sennrich2015neural}.

Dictionary coder algorithms like BPE learn dictionaries of tokens with the purpose of representing strings with as few tokens as possible, increasing the level of compression. This reduces the effective depth of the unrolled RNN network (i.e. the shortest path through the graph in figure \ref{fig:multimc}), which is a reasonable learning objective for our dictionary.

\begin{algorithm}
\caption{Adapted byte-pair encoding algorithm}
\begin{algorithmic}
  \REQUIRE $T_{max}, T = \{s_1, \dots, s_m\}, \bm s = s_{i_1}, \dots, s_{i_n}$
  \COMMENT{Initial dictionary $T = \Sigma$, string $\bm s$}
  \WHILE{true}
    \STATE $j, k \mapsfrom \argmax_{j,k} paircount(s_j, s_k)$
    \STATE $s_{\text{new}} \mapsfrom [s_j | s_k],\quad T \mapsfrom T \bigcup \{s_{\text{new}}\}$
        \IF{$|T| = T_{max}$}
       \STATE $\textbf{break}$
    \ENDIF
    \STATE Substitute each occurrence of $s_j,s_k$ in $\bm s$ with $s_{\text{new}}$
    \FORALL{$l \in \{l : count(s_l) < count(s_{new})\}$}
       \STATE $T \mapsfrom T \setminus \{s_l\}$
       \STATE Substitute each occurrence of $s_l$ in $\bm{s}$ with $s_o,s_p$ s.t. $[s_o | s_p] = s_l$
    \ENDFOR
  \ENDWHILE
  \STATE \textbf{return} $T$
\end{algorithmic}
\end{algorithm}

Regular BPE starts with a dictionary of characters and consecutively replaces the most frequent pairs of tokens with a single new token, until a given dictionary size $T_{max}$ is reached. We extend the algorithm by reversing the merger of two tokens whenever a token becomes too rare. As a motivating example consider the string \emph{abcabcabc$\dots$}. In this case \emph{a} and \emph{b} are merged into \emph{ab}, followed by a merger between \emph{ab} and \emph{c} into \emph{abc}. Our extension makes sure that the token \emph{ab}, which now occurs zero times, is removed from the dictionary. This removal prevents us from wasting space in the dictionary on rare tokens.

Our implementation of this algorithm uses a bit array of the size of the input data, where each element signifies whether the corresponding character is merged with the subsequent character. We maintain two $d$-ary heaps of tokens and token pairs sorted by their frequency. The algorithm proceeds by repeatedly popping the most common pair from the heap and searching the text and bit array for occurences. If an occurence is found, the bit array is updated to represent the merge, and the $d$-ary heaps are updated to reflect the new token and pair counts. This requires a minimum of $D$ passes over the data to construct a dictionary of size $D$ but uses a relatively small amount of memory.

\section{Experiments}

\begin{table}
  \centering
  \small
  \begin{tabular}{rl}
    \hline \noalign{\smallskip}
    {\bf \#} & {\bf Token}\\\noalign{\smallskip}\hline
    1 & and$\cdot$ \\
    2 & a$\cdot$ \\
    3 & the$\cdot$ \\
    7 & s$\cdot$ \\
    8 & of$\cdot$the$\cdot$ \\
    11 & in$\cdot$the$\cdot$ \\
    12 & ed$\cdot$ \\
    17 & ing$\cdot$ \\
    65 & ation$\cdot$ \\
    239 & people$\cdot$ \\
    245 & man \\
    296 & ed$\cdot$by$\cdot$the$\cdot$ \\
    525 & external$\cdot$links$\cdot$ \\
    540 & at$\cdot$the$\cdot$end$\cdot$of$\cdot$the$\cdot$ \\
    565 & at$\cdot$the$\cdot$university$\cdot$of$\cdot$ \\
    608 & united$\cdot$states$\cdot$ \\
    1468 & in$\cdot$the$\cdot$united$\cdot$states$\cdot$ \\
    2727 & one$\cdot$of$\cdot$the$\cdot$most$\cdot$\\\hline
  \end{tabular}
  \caption{
A sample from the tokens in the dictionary of size 8192 constructed using the text8 dataset by our adapted BPE algorithm. Spaces are visualized with the $\cdot$ character. The most common tokens are similar to the ones traditionally found in word-level models e.g. `and ', `the ', and `a '. The dictionary also contains common suffixes such as `s ' (for plural nouns and third person singular verbs), `ing ' (for gerunds and verbal actions), and `ed ' (for adjectives, past tenses and past participles), as well as multi-word tokens e.g. `of the', `and the', `in the', etc. and longer phrases.
}
\end{table}

\subsection{Implementation}

The irregular, data-dependent access pattern makes the multiscale model difficult to implement in a performant manner using existing GPU-accelerated deep learning frameworks such as Theano and Torch. Hence, experiments were performed with a hand-written CUDA implementation of both the model (including layer normalization) and the dynamic programming forward and backward sweeps. Our implementation was able to exploit the parallelism inherent in the model, fully utilizing the K20 and K80 NVidia GPUs that the models were trained on.

\begin{figure}
\begin{tikzpicture}[scale=0.8]
    \begin{axis}[
    	title = {Training curves},  
    	xlabel = {Updates},
    	ylabel = {Bits per character},
      xmin = 0, xmax = 5000,
    	ymin = 1, ymax = 3,
    	ytick = {1,1.5,2,2.5,3},
    	xtick = {0,1000,...,5000},
			minor y tick num={3},
			minor x tick num={1},
      legend entries = {Regular LSTM,Multiscale LSTM},
      legend pos=north east
    	]
    	\addplot[mark=none,dashed] table {char.csv};
    	\addplot[mark=none] table {multiscale.csv};
    \end{axis}
\end{tikzpicture}
\caption{Training curves of the regular and multiscale LSTM. Note how the multiscale LSTM training loss starts lower because of the learned dictionary, which shows that the use of compression algorithms for dictionary construction is effective.}
\end{figure}
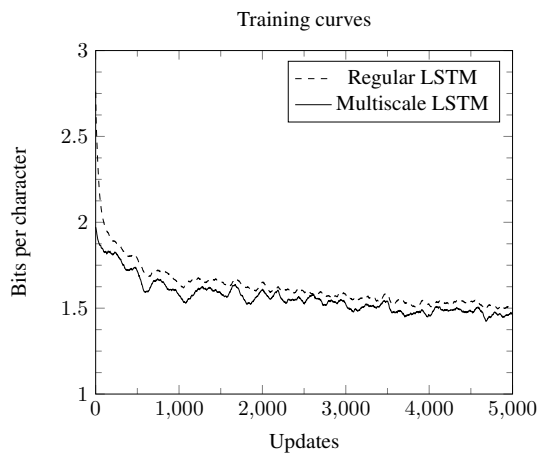

\subsection{Penn Treebank}

\begin{table*}
  \centering
  \small
  \begin{tabular}{l}
    \hline \noalign{\smallskip}
    {\bf Samples} \\ \noalign{\smallskip} \hline \noalign{\smallskip}
    the $\cdot$independ$\cdot$ence $\cdot$in the $\cdot$third quarter $\cdot$the $\cdot$chief $\cdot$ex$\cdot$port $\cdot$stock $\cdot$prices $\cdot$for the $\cdot$year$\cdot$ $\cdot$and $\cdot$into the $\cdot$disa$\cdot$ster$\cdot$l$\cdot$and\\
    gains $\cdot$so $\cdot$on the $\cdot$economy $\cdot$because $\cdot$in addition $\cdot$to a $\cdot$compl$\cdot$ex $\cdot$closed $\cdot$higher $\cdot$comm$\cdot$ut$\cdot$e $\cdot$pres$\cdot$sure $\cdot$of $\cdot$his $\cdot$company\\
    meeting $\cdot$in $\cdot$the $\cdot$trust $\cdot$is $\cdot$expected to $\cdot$be $\cdot$an$\cdot$ticip$\cdot$ated $\cdot$\$$\cdot$ $\cdot$offic$\cdot$es $\cdot$during the $\cdot$past \\
    actu$\cdot$ally $\cdot$have $\cdot$spok$\cdot$es$\cdot$man $\cdot$with $\cdot$hous$\cdot$ing $\cdot$their $\cdot$junk $\cdot$bond $\cdot$due $\cdot$\$ N billion from $\cdot$most $\cdot$important $\cdot$next $\cdot$day $\cdot$at N\\\hline
  \end{tabular}
  \caption{Samples from the multiscale model trained on Penn Treebank. Token boundaries are marked with the $\cdot$ symbol. The samples show the model's ability to model a sentence by predicting entire words or phrases (`in addition', `into the') at a time, while also being able to exploit subword structure (`comm$\cdot$ut$\cdot$e') and maintaing the flexibility of character language models to output unseen words (`disa$\cdot$ster$\cdot$l$\cdot$and').}
\end{table*}

We evaluate the multiscale model on the widely used Penn Treebank dataset using the training, validation and test split proposed by Mikolov~\cite{mikolov2012subword}. Our baseline is an LSTM with 1024 hidden units and embeddings of dimension 512 trained with truncated backpropagation-through-time (TBPTT) on sequences of length 400. These optimal values were found using a grid search. We train using the Adam~\cite{kingma2014adam} optimizer (learning rate of 0.001) and to increase convergence speed we use layer normalization~\cite{ba2016layer}.

Our baseline model achieves a score of 1.43 bits per character. The multiscale model is trained using the exact same configuration but using a dictionary of 2048 tokens. It achieves a test score of 1.42 bits per character. Note that the multiscale models improvements are orthogonal to what can be achieved by straightforwardly increasing the capacity of the network. The regular LSTM networks with more than 1024 units showed decreased performance in our experiments due to overfitting.

Moreover, our network is able to achieve better performance with far fewer parameters. The multiscale model with 512 hidden units, embeddings of size 256, and 2048 tokens has 51\% fewer parameters compared to our baseline, but achieves a score of 1.41 bpc, compared to 1.48 bpc for a regular LSTM with the same embedding and hidden state size.

\subsection{Text8}

Text8~\cite{mahoney2009large} is a text dataset of 100 million characters built from the English Wikipedia. The characters are limited to the 26-letter alphabet and spaces. We use the traditional split of 90, 5 and 5 million for the training, validation and test set respectively.

We compare the performance of our multi-scale model with a single-layer character-level language model with 2048 units. The same training procedure as described in the previous subsection is used. The baseline achieves a score of 1.45 bits per character. The multiscale model improves on this performance using a dictionary of 16,384 tokens, achieving a test score of 1.41 bits per character.

\section{Related work}

In language modeling a variety of approaches have attempted to bridge the gap between character and word-level models. The approach in \cite{kim2015character} is to apply a convolutional neural network (CNN) with temporal pooling over the constituent characters of a word. The CNN filters of this network can be interpreted as character n-gram detectors. The output of this network is used as the input to an LSTM network which models the word-level dynamics. Note that the resulting model still requires information about word-boundaries.

Other approaches use multi-scale RNN architectures. The model in~\cite{bojanowski2015alternative} uses both a word-level and character-level RNN, the latter being conditioned on the former. This model too still requires knowledge of word boundaries. The approach in~\cite{chung2016hierarchical} does not require word boundaries, and instead uses the straight-through estimator to learn the latent hierarchical structure directly. Their model does not learn separate embeddings for the segments however, and can only output a single character at a time.

The latent sequence decomposition (LSD) model introduced in~\cite{chan2016latent} is related to our multiscale model, and was shown to improve performance on a speech recognition task. Instead of using compression algorithms the LSD model uses a dictionary of all possible $n$-grams. Since the number of $n$-grams grows exponentially, this limits the the dictionary to very short tokens only. The LSD model uses a regular RNN which is trained on a set of sampled segmentations instead of averaging the hidden states using dynamic programming. This complicates training and makes the likelihood of the model intractable. The recent Gram-CTC model~\cite{liu2017gram} is also related and does use dynamic programming but still uses a dictionary of character n-grams.

Although our model is competitive with recent methods such as MI-LSTM~\cite{wu2016multiplicative} and td-LSTM~\cite{zhang-lu-lapata:2016:N16-1}, which achieve 1.44 and 1.63 bits per character on the text8 dataset respectively, other recent models such as HM-LSTM~\cite{chung2016hierarchical} have achieved lower scores (1.29 bpc). Since many of the LSTM variations in the literature can be extended to the multiscale model, we believe it is possible to improve the performance of multiscale models further in the future. Similarly, deeper multi-layer extensions to our model are feasible.

\section{Discussion}

Through arithmetic encoding it can be shown that modeling data is equivalent to compressing it~\cite{mahoney1999text}. Using neural networks to improve upon text compression algorithms is a common technique~\cite{mahoney2000fast}, but as far as we are aware the reverse has not been researched. One can see our model as a mix between non-parametric and parametric approaches: As discussed in section \ref{sec:compositionality}, character-level models learn a parametric mapping from constituent characters to semantic representations of morphemes. Word-level models avoid learning this highly non-linear function by constructing a dictionary and learning a representation for each word, which is non-parametric. Our multiscale model generalizes this approach, combining non-parametric dictionary coders and parametric RNN models. The size of the dictionary allows us to choose the balance between the two approaches.

A rough parallel can be drawn between our multiscale approach for sequences and superpixels in the computer vision domain~\cite{ren2003learning}, where pixels are clustered in order to improve computational and representational efficiency

The multiscale model can also be related to work on text segmentation. The hierarchical Pitman-Yor language model in~\cite{mochihashi2009bayesian} learns how to segment a string of characters into words, while simultaneously learning a word-level $n$-gram model. Each path through the graph of the multiscale model (figure \ref{fig:multimc}) can be considered a single segmentation of the text, with the likelihood of the string being the marginalization over all possible segmentations.

A large number of combinations of data compression and neural network sequence modeling are still open to investigation. Besides BPE, there are many other dictionary coder algorithms out there. Another consideration would be to learn the dictionary and the sequence model jointly. Subsequently, a variety of neural network models can conceivably be adapted to work with the multi-scale representation of text used in this paper e.g. bag-of-words (BOW) models could be replaced with bag-of-token models instead, similar to the approach in~\cite{bojanowski2016enriching} which uses character $n$-grams.




\bibliography{aigaion,ml,nn,predoc,own,nips2015}
\bibliographystyle{emnlp}

\end{document}